\documentclass{article}
\usepackage{spconf,amsmath,graphicx}
\usepackage{booktabs} 
\usepackage{multirow}
\usepackage{epsfig}
\usepackage{bm}
\usepackage{amsmath}
\usepackage{amssymb}
\usepackage{diagbox}
\usepackage{caption}

\usepackage[labelsep=endash]{caption}
\title{PGTRNet: Two-phase Weakly Supervised Object Detection with Pseudo Ground Truth Refinement}
\name{Jun Wang$^1$, Hefeng Zhou$^{2}$,Xiaohan Yu$^3$}
\address{$^1$University of Warwick $^2$Shanghai Jiao Tong University $^3$Griffith University  \\
jun.wang.3@warwick.ac.uk, zadeji1@sjtu.edu.cn, xiaohan.yu@griffith.edu.au}

\begin{document}
\ninept
\maketitle
\begin{abstract}
 Current state-of-the-art weakly supervised object detection (WSOD) studies mainly follow a two-stage training strategy which integrates a fully supervised detector (FSD) with a pure WSOD model. There are two main problems hindering the performance of the two-phase WSOD approaches, i.e., insufficient learning problem and strict reliance between the FSD and the pseudo ground truth (PGT) generated by the WSOD model. This paper proposes pseudo ground truth refinement network (PGTRNet), a simple yet effective method without introducing any extra learnable parameters, to cope with these problems. PGTRNet utilizes multiple bounding boxes to establish the PGT, mitigating the insufficient learning problem. Besides, we propose a novel online PGT refinement approach to steadily improve the quality of PGT by fully taking advantage of the power of FSD during the second-phase training, decoupling the first and second-phase models. Elaborate experiments are conducted on the PASCAL VOC 2007 benchmark to verify the effectiveness of our methods. Experimental results demonstrate that PGTRNet boosts the backbone model by 2.1\% mAP and achieves the state-of-the-art performance. 
\end{abstract}
\begin{keywords}
Weakly Supervised Object Detection, Pseudo Ground Truth, Online Refinement
\end{keywords}
\section{Introduction}
\label{sec:intro}

Object detection aims to detect and localize different semantic instances in images or videos with bounding boxes and obtain their categories. It is a significant sub-field of Computer Vision and has been extensively applied to many fields such as autonomous vehicles, security monitoring, and robotics. However, there are complex challenges since one image may contain different classes of objects and these instances may occur in different parts of images or videos. \par
Owing to the flourish of Deep Learning (DL) techniques, various DL-based algorithms have been proposed in object detection \cite{girshick2014rich,girshick2015fast,redmon2016you,ren2020instance}. Despite a great success has been achieved in these methods, they often require a large number of box-level annotations to improve their accuracy and generalization ability. This requirement, on the other hand, cannot always meet in real world applications. Therefore, Weakly Supervised Object Detection (WSOD), aiming to perform object detection with only image-level annotation, has gained increasing research attention. Most previous works \cite{bilen2016weakly,wang2015relaxed,cinbis2016weakly,tang2017deep} exploit the Multiple Instance Learning (MIL) strategy for WSOD. Moreover, architectures combining the MIL and CNN demonstrate better performance due to the significant feature extraction ability of CNN compared with traditional hand-designed features. Research has shown that an end-to-end training or its variant can further improve the results for WSOD. 
Recent powerful WSOD methods normally follow a two-phase learning approach. A MIL detector combined with the CNN feature extractor are trained with image-level labels in the first phase. Outputs of the first-phase model are then taken as the pseudo ground truth (PGT) to fine-tune a Fully-Supervised Object Detection (FSOD) detector, e.g., Fast r-cnn \cite{girshick2015fast}. \par
However, most of previous works focus on the first-phase WSOD model and almost no study explores the potential of the second-phase training. We observe that current two-phase WSOD models regard the top one bounding box as the PGT, but the top one bounding box may only contain a part of instances or objects, and thus, may confuse the detectors. We denote this as the insufficient learning problem. In addition, PGT does not change during the whole training of the second-phase detector, hence even worse results will be obtained given the inaccurate PGT. To cope with these problems, we propose a pseudo ground truth refinement network (PGTRNet) which takes advantage of multiple bounding boxes to produce the PGT. Furthermore, PGTRNet gradually refines the PGT to release the reliance of the WSOD detector. \par

To sum up, this work mainly has four contributions. \textbf{(1)} We propose a novel two-phase WSOD architecture, the PGTRNet, which takes advantage of PGT by the first-phase detector training and gradually refines them during the second-phase training, alleviating the reliance between the WSOD and FSOD models. Moreover, PGTRNet is simple yet effective without introducing any extra learnable parameters. \textbf{(2)} Different from previous works that only regard the top one bounding box as PGT, we propose to consider more bounding boxes to generate the PGT, mitigating the insufficient learning problem. \textbf{(3)} We develop multiple strategies to effectively refine the PGT during the training of the second-phase detector. \textbf{(4)} We verify the effectiveness of our PGTRNet on the VOC2007 benchmark. Experimental results demonstrates that PGTRNet improves the backbone network by 2.1\% mAP and achieves state-of-the-art performance. Further ablation studies show the necessity of PGT refinement in the FSOD detector training. 

\section{Related Work}
\label{sec:literature}

The main issue of WSOD is that it is much harder to generate the bounding box of objects as no ground truth location is available. In WSOD, the model may be misled and this may be amplified if the given supervision is always inaccurate. To solve the WSOD problem, current methods normally adopt the Multiple Instance Learning (MIL) \cite{prest2011weakly} strategy and take advantage of CNN, and form an end-to-end unified learning approach.  \par
Weakly Supervised Deep Detection Network (WSDDN), proposed by Bilen \& Vedaldi  \cite{bilen2016weakly} is a two-stream architecture. One stream for classification and one stream for detection. The positive sample is determined in terms of the score of these streams. Based on WSDDN, Kantorov et al. \cite{kantorov2016contextlocnet} introduces the use of context (either additive or contrastive) to aid in localization. This is also in accordance with the spirit of recent research \cite{wang2021mask, YUICCV21,YU2021108067,yu2020patchy}, which focuses on localizing vital regions. One major problem of these methods is that the algorithm tends to get trapped in a local minimum.. To alleviate this problem, Tang et al. \cite{tang2017multiple} presents an online instance classifier refinement (OICR) algorithm which is also based on WSDDN. Also proposed by Tang et.al. \cite{tang2018pcl}, Proposal Cluster Learning (PCL) clusters the region proposals of an image into different clusters before sending them as the inputs to the backbone network, making the algorithm pay attention to the whole part of object. \par
Recent state-of-the-art WSOD approaches follow a two-phase learning strategy. A weakly supervised network (e.g., methods mentioned above) is first trained to obtain the predicted bounding boxes on a training dataset. These predicted bounding boxes are then regarded as the PGT to fine-tune a fully supervised detector. Most approaches \cite{tang2018pcl,tang2017multiple,zhang2018ml,tang2018weakly} exploit Fast r-cnn as the second-phase network and outperform the pure WSOD model. However, as mentioned in the previous section, insufficient learning problem limits the performance of the these methods. Besides, the effectiveness of the fully-supervised detector is highly dependent on the quality of the PGT generated by the WSOD detector.

\begin{figure*}[ht]   
    \centering 
    \includegraphics[width=0.71\textwidth]{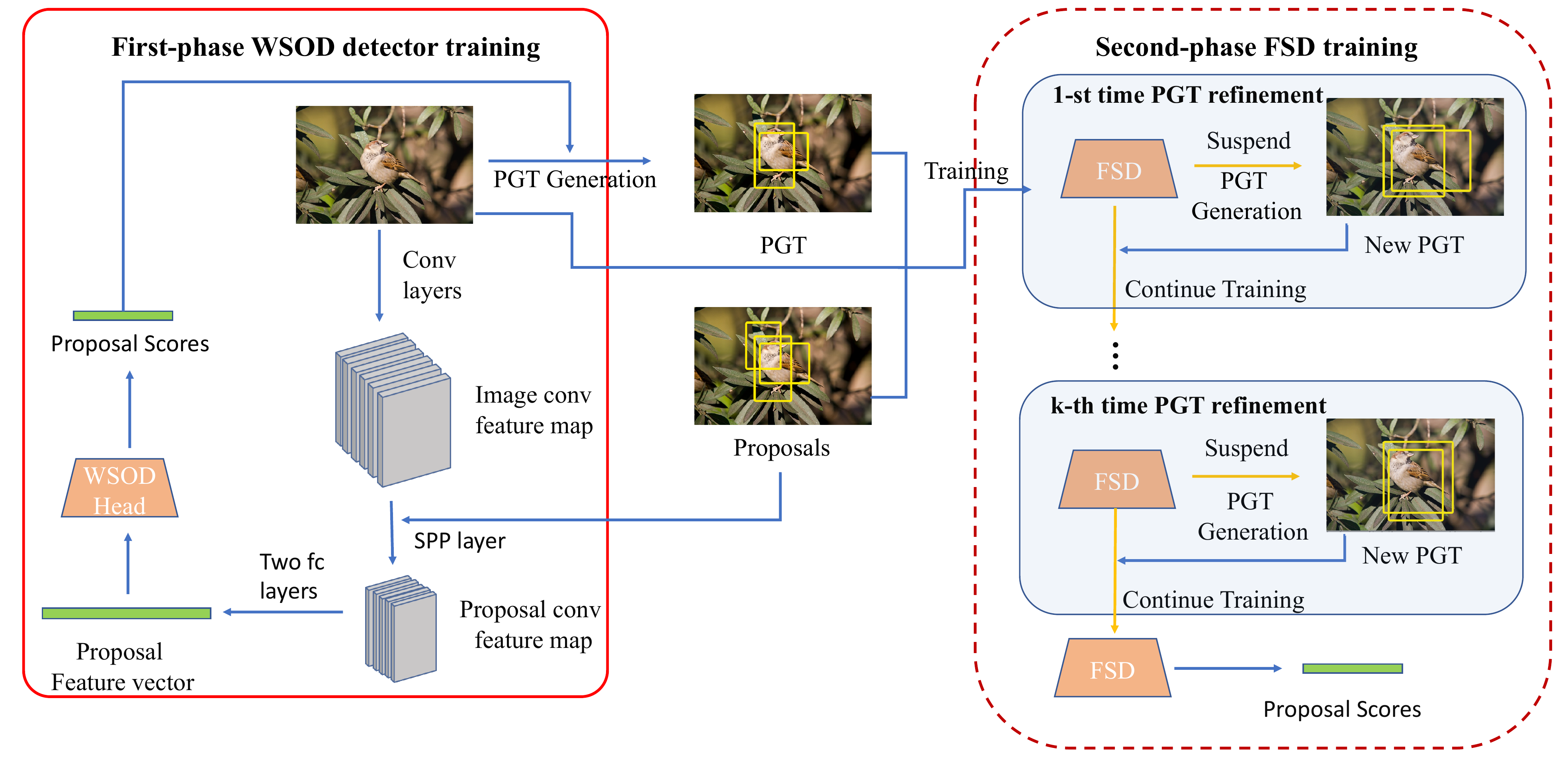} 
    \caption{The general architecture of the PGTRNet. We first train the first-phase WSOD detector according to the training images and the corresponding region proposals. Inference is then applied to these images to obtain the proposal scores. PGTRNet then produces the PGT via the proposed PGT generation method. The PGT is then utilized to supervised the learning of the fully supervised detector (FSD). To improve the quality of the PGT and alleviate the reliance between the first and second-phase detector, PGTRNet refines the PGT multiple times based on the outputs of the FSD and the refinement strategies.}    
    \label{architecture}
\end{figure*}

\section{Methods}
\label{sec:methods}

This section describes our proposed methods to advance the precision of WSOD. PGTRNet adopts the PCL \cite{tang2018pcl} and the Faster r-cnn \cite{ren2015faster} as the backbone networks of the first-phase and second-phase detectors, respectively. We first briefly review the general framework of the two-phase WSOD model. Our PGTRNet can be built on top of any two-phase WSOD architecture. Figure 1 illustrates the general architecture of PGTRNet.  Images and the associated region proposals are fed into the first-phase WSOD detector to train the detector and obtain the proposal scores $S_c$. PGT generation is then applied to produce the PGT according to the region proposals and their corresponding scores. Combined with the images, PGT is then sent to the fully supervised detector (FSD) to supervise the learning. During the training of the FSD, we suspend the training and apply the PGT refinement multiple times to improve the quality of PGT based on the refinement strategy. The details of PGTRNet including PGT generation and PGT refinement are described in section 3.2.

\subsection{Two-phase WSOD}
We denote a training image as $ \bm{I}\subseteq\bm{R}^{H\times W \times 3}$ and its image-level annotation as $ \bm{y} = [y_1,y_2,...,y_C]\subseteq{[0,1]^C}$. Note that $H$ and $W$ are the height and width of the image. $C$ is the total number of classes. Given the $\bm{I}$ and $\bm{y}$, WSOD aims to predict the potential objects (bounding boxes) $\bm{B}_c$ for class $c$ and corresponding scores $\bm{S}_c$. $N$ is the total number of the outputs from the WSOD model.
\begin{equation}
 \bm{B}_c=\{b_{c,1}, b_{c,2}, ..., b_{c,N}\}
\end{equation}

\begin{equation}
 \bm{S}_c=\{s_{c,1},s_{c,2}\,...,s_{c,N}\}
\end{equation}

Current state-of-the-art WSOD methods often follow a two-phase training strategy. Specifically, in the first stage, a pure WSOD model is trained with only $\bm{I}$ and $\bm{y}$. Then, training images $\bm{I}$ are fed into the trained WSOD model again to obtain the predicted bounding boxes $\bm{B}_c$ for class $c$. After obtaining $\bm{B}_c$, top one bounding box $\bm{b^{*}_c}$ is selected as the pseudo ground truth to fine-tune a fully-supervised detector. Equations (3) shows this procedure.
\begin{equation}
 \bm{k}=argmax_{i}(s_{c,i}),  \quad \bm{b}^*_c=b_{c,k} 
\end{equation}

%[PCL] has proven that a Fast RCNN can boost the performance of their ensembling method. This paper adopts the representative WSOD model PCL [] as the first-phase model and [] [] as the fully supervised detector. 
%To implement our idea, a PCL network is first trained with only image-level annotations. This means that bounding boxes of each instance and the number of instances for each class in the image are not available during the training of PCL. Hence, algorithms only have the knowledge about what categories exist in the image. Built in the trained PCL model, the algorithm then performs inference on the training dataset to obtain the predicted bounding boxes and their categories. Next, we establish instance-level annotations to train the second-phase model according to the results obtained in the previous step. During the training on the second-phase network, algorithms know the number of instances in each class and their coordinates information in images, but this information is created by the first-phase network. Hence, it is the PGT. After training, the performance of this two-phase WSOD architecture will be verified through testing the second-phase detector on the test dataset. One significant procedure in this architecture is how to establish the PGT for the second-phase network from the the first-phase model.

\subsection{PGTRNet}
\subsubsection{PGT Generation}
PGT is the core to connect the first-phase and second-phase models. As noted above, most previous works select the top one bounding boxes as the PGT to train a second-phase model such as Fast r-cnn. It is very natural to select the  highest confidence bounding box since detectors tend to be trained with correct supervision according to these high score boxes. Nonetheless, this could lead to some problems hindering the performance of the detector. We empirically observe that objects from the same class may appear more than once in a large number of pictures and datasets. In this case, only given the top one bounding boxes, the detector may be unable to learn adequate features to distinguish this class from others. This problem could be exacerbated when the dataset is small or there are many subcategories in a class such as Border Collie and Akita in the dog category. Besides, it is likely that the top bounding boxes are misclassified or contain only parts of an object in WSOD. These cases may also cause an insufficient learning problem.  \par

To mitigate the problems mentioned above, we propose to consider top $k$ bounding boxes as the PGT to guarantee a high proposal recall. Nonetheless, a large $k$ value may also introduce many negative samples, and thus mislead the learning while a small $k$ value may lose significant learning information. Hence, it is essential to maintain the balance between high proposal recall and high precision. We collect top $k$ bounding boxes in three steps. \par
(1) We train a WSOD model according to the images and associated image-level annotation, and then perform inference on the same dataset to obtain the predicted bounding boxes and their scores. \par
(2) Recall that algorithms do not know the number of instances in each category in images and their bounding boxes but they do have knowledge about which class occurs in the image for WSOD. We record this information via processing the annotation file. \par
(3) Based on information obtained from step one and two, top $k$ bounding boxes $\bm{B}^{'}_{c}$ for class $c$ are created by ordering the bounding boxes based on their scores $\bm{S}_c$ in corresponding categories and selecting the best $k$ boxes. We then save the PGT to establish an instance-level annotation file which will be processed to train the second-phase model.
Equations (3) then changes to be Equations (4) and (5).
\begin{equation}
 \bm{L}_c=argsort(\bm{B}_c)[1\sim k]
\end{equation}

\begin{equation}
 \bm{B}^{'}_c={\{b_{c,1}, b_{c,2},..,b_{c,i},..,b_{c,k}\}}, i\subseteq \bm{L}_c
\end{equation}

\subsubsection{PGT Refinement}
Previous works often focus on designing decent pure WSOD model and neglect the second-phase fine-tuning procedure. We argue that the second-phase training has huge potential to improve the results of the WSOD model. Given the PGT, it has been shown that a fully supervised model may improve the results from a pure WSOD model. However, for those misclassified samples and samples only containing non-distinguishable parts, a second-phase detector seems powerless. Considering this, we propose to refine the PGT during the training. This strategy can improve the PGT or increase the number of positive training samples, and thus may boost the performance. 
\par
PGTRNet performs inference on the training dataset during the training of the fully supervised detector and recreates instance-level annotations according to the bounding box scores $S_c$. Unlike the PGT generation in Section 3.2.1, this update is performed online. This means that the weights and gradients of the network are maintained after performing inference on training dataset. There are two reasons for choosing this strategy. \textbf{(1)} Compared with completing training and then refining the PGT in a new model, this strategy has lower computational complexity. \textbf{(2)} During the training, the detector may not be misled from those negative training samples much. \par
Then, the problem turns to how and when to refine the PGT. We propose seven strategies to refine the PGT. These strategies are divided into two groups as shown in Table 1 Strategies in group one determine the time to refine the PGT, while those in group two define how to update the PGT. A complete refinement strategy comprises one strategy from group one and one strategy from group two. 

To be specific, we do PGT refinement in the following steps. As a necessary procedure, a PCL model is trained to generate the annotations (PGT) for the second-phase detector via the method described in Section 3.2.1. We then train the second-phase model based on the PGT. At specific epochs during the training (e.g., epoch mentioned in group one), we suspend training, save the model, and change the model from training mode to evaluation mode. After suspending the training, we perform inference on training images and then update the annotation files according to the strategies mentioned in group two. One thing that should be emphasized is that we do not stop the training, and thus, the gradient and the weight of the network are kept. The next step is to continue training according to the refined annotations. 

\begin{table}[htbp]
\scriptsize
	\centering  
	\caption{Group one and two of the PGT refinement strategies.}

	\begin{tabular}{l}  
	\toprule
 \textbf{Group 1}: \\
 \hline
 	1. Refine the PGT after each epoch. \\
 	2. Refine the PGT every three epochs. \\
 	3. Refine the PGT after each epoch in the last three epochs. \\
	4. Refine PGT only once at the 2/3 of the maximum epochs. \\
	\toprule
 \textbf{Group 2}: \\
 \hline
	1. Update all k PGT . \\
	2. Only update half of the k best PGT. \\
	3. Only update half of the k worst PGT. \\

\bottomrule
	\end{tabular}
\end{table}

\begin{table*}[ht]
\scriptsize
\caption{The comparisons to state-of-the-art methods on the test set of PASCAL VOC 2007. The last line is our proposed method. The best AP of each class and mAP are highlighted.}
\centering
\begin{tabular}{c|p{0.1cm}<{\centering} p{0.1cm}<{\centering} p{0.1cm}<{\centering} p{0.1cm}<{\centering} p{0.2cm}<{\centering} p{0.1cm}<{\centering} p{0.1cm}<{\centering} p{0.1cm}<{\centering} p{0.2cm}<{\centering} p{0.1cm}<{\centering} p{0.2cm}<{\centering} p{0.1cm}<{\centering} p{0.2cm}<{\centering} p{0.3cm}<{\centering} p{0.3cm}<{\centering} p{0.2cm}<{\centering} p{0.2cm}<{\centering} p{0.15cm}<{\centering} p{0.15cm}<{\centering} p{0.3cm}<{\centering}|p{0.4cm}<{\centering}}
\toprule  
Method& aero& bike& bird& boat& bottle& bus& car& cat& chair& cow& table& dog& horse& mobile& person& plant& sheep& sofa& train& tv& mAP \\
\midrule  
PCL-Ens+FRCNN \cite{tang2018pcl} &63.2 &69.9 &47.9 &22.6 &27.3 &71.0 &69.1 &49.6 &12.0 &60.1
&51.5 &37.3 &63.3 &63.9 &15.8 &23.6 &48.8 &55.3 &61.2 &62.1&48.8 \\
\hline

ML-LocNet-L+ \cite{kantorov2016contextlocnet} &60.8 &70.6 &47.8 &30.2 &24.8 &64.9 &68.4 &57.9 &11.0 &51.3 &55.5 &48.1 &68.7 &69.5 &28.3 &25.2 &51.3 &56.5 &60.0 &43.3&49.7 \\
\hline

WSRPN-Ens+FRCNN \cite{tang2018weakly} &63.0 &69.7 &40.8 &11.6 &27.7 &70.5 &74.1 &58.5 &10.0 &66.7
&\textbf{60.6} &34.7 &\textbf{75.7} &70.3 &25.7 &26.5 &55.4 &56.4 &55.5 &54.9&50.4 \\
\hline

Multi-Evidence \cite{ge2018multi} &64.3 &68.0 &56.2 &36.4 &23.1 &68.5 &67.2 &64.9 &7.1 &54.1 &47.0 &57.0 &69.3 &65.4 &20.8 &23.2 &50.7 &59.6 &65.2 &57.0&51.2 \\
\hline

W2F+RPN+FSD2 \cite{zhang2018w2f} &63.5 &70.1 &50.5 &31.9 &14.4 &72.0 &67.8 &73.7 &23.3 &53.4
&49.4 &65.9 &57.2 &67.2 &27.6 &23.8 &51.8 &58.7 &64.0 &62.3&52.4 \\
\hline

OIM+IR+FRCNN \cite{lin2020object} &53.4 &72.0 &51.4 &26.0 &27.7 &69.8 &69.7 &74.8 &21.4 &67.1
&45.7 &63.7 &63.7 &67.4 &10.9 &25.3 &53.5 &60.4 &70.8 &58.1&52.6 \\
\hline

PredNet-Ens \cite{arun2019dissimilarity}&\textbf{67.7} &70.4 &52.9 &31.3 &26.1 &\textbf{75.5} &73.7 &68.6 &14.9 &54.0
&47.3 &53.7 &70.8 &70.2 &19.7 &29.2 &54.9 &\textbf{61.3} &67.6 &61.2&53.6 \\
\hline

WSOD2 \cite{zeng2019wsod2} &65.1 &64.8 &57.2 &\textbf{39.2} &24.3 &69.8 &66.2 &61.0 &\textbf{29.8} &64.6
&42.5 &60.1 &71.2 &70.7 &21.9 &28.1 &58.6 &59.7 &52.2 &64.8&53.6 \\
\hline

GAM+PCL+Reg \cite{yang2019towards} &59.8 &72.8 &54.4 &35.6 &30.2 &74.4 &70.6 &\textbf{74.5} &27.7 &68.0
&51.7 &46.3 &63.7 &68.6 &14.8 &27.8 &54.9 &60.9 &65.1 &67.4&54.5 \\
\hline

Wetectron \cite{ren2020instance} &68.8 &\textbf{77.7} &57.0 &27.7 &28.9 &69.1 &\textbf{74.5} &67.0 &32.1 &\textbf{73.2} &48.1 &45.2 &54.4 &\textbf{73.7} &\textbf{35.0} &29.3 &\textbf{64.1} &53.8 &65.3 &65.2&54.9 \\
\hline

\textbf{PGTRNet (ours)} &65.6 &73.0 &\textbf{61.2} &29.9 &\textbf{34.4} &69.0 &67.9 &60.7 &28.8 &67.7
&52.9 &\textbf{71.6} &68.5 &67.2 &12.3 &\textbf{30.9} &43.2 &59.1 &\textbf{72.1} &\textbf{70.0}&\textbf{55.3} \\

\bottomrule 
\end{tabular}
\end{table*}

\section{Experiments}
We verify the effectiveness of PGTRNet on one of the most common WSOD benchmark PASCOL VOC 2007 dataset. Mean Average Precision (mAP) is used to evaluate the model. Section 4.1 shows the details to reproduce our work. Experimental results and comparisons to SOTA are given in the following sections. Eventually, an ablation study is demonstrated to further prove the effectiveness of our propose methods. 
\subsection{Implement Details \& Dataset}
Following the most previous two-phase works, we use five image scales \{480, 576, 688, 864, 1200\} to enable multi-scale training and testing via resizing the shortest side to the closest scales. $k$ in equations (4) and (5) is set to 2. We train all models with 12K iterations and the minibatch size is set to 4. The initial learning rate is set to 0.002 and decayed in 8000 and 10500 iterations. All experiments are conducted on one Nvidia GTX 2080Ti GPU card.
\par
The PASCAL VOC 2007 \cite{everingham2008pascal} dataset is one of the most common benchmarks to measure the ability of a model in object detection. A large amount of works test their models and report the results on this dataset. This benchmark consists of 5011 training images and 4952 testing images with a total amount of 20 categories.

\begin{table}[!htbp]
\centering
\scriptsize
\caption{The mAP of different combination of strategies of PGTRNet on the test set of PASCAL VOC 2007 benchmark. $-$ (no strategy been utilized, or no results). The best mAP is highlighted in bold.}
\begin{tabular}{c|c|c|c|c}
\toprule  
\diagbox{GP1}{mAP}{GP2}&$-$&1&2&3\\ 
\midrule 
$-$&53.54& $-$&$-$ &$-$\\
\hline
1& $-$& 27.85& 55.14& 54.44\\
\hline
2& $-$& 45.88& 54.33& \textbf{55.29}\\
\hline
3& $-$& 45.31& 53.78& 54.59\\
\hline
4& $-$& 54.56& 54.00& 53.87\\
\bottomrule
\end{tabular}
\end{table}

\subsection{Detection Results}
Here, we demonstrate the detection results of PGTRNet and comparisons to state-of-the-art methods. Table 2 gives comparisons of PGTRNet to recent state-of-the-art methods on PASCAL VOC 2007 dataset. Obviously, PGTRNet beats all methods with a mAP of 55.3\% and outperforms the second best-performed method by 0.4\%. Moreover, PGTRNet achieves the best average precision on almost a third of all categories (6 out of 20). These encouraging results confirm the superiority of PGTRNet. Table 3 shows the performance of PGTRNet with different combination of PGT refinement strategies from groups one and two. We denote a complete strategy as */* and * is the $*^{th}$ strategy of group1/2. Note that the proposed PGT generation is added on all experiments including the group $-/-$. 
It can be observed that strategies 2 and 3 of group two combining with any strategy from the group one boost the performance by almost 1\% mAP. These strategies only updates half of the PGT, preventing the network from significant fluctuation while refining the PGT. The best strategy 1/2 improves the mAP from 53.54\% to 55.29\% (1.74\% mAP) which achieves the state-of-the-art performance. \par
However, strategies {1,2,3}/1 noticeably limit the performance possibly due to the fluctuation of the network. Strategy 1 of group two updates all the PGT, and the model fails to converge if the algorithm frequently performs refinement. A different result can be seen in strategy 4/1 which boosts the mAP by 1\%.  A possible reason is that the algorithm performs refinement only once in strategy 4 of group one, and the detector still has adequate training epochs to fit the updated PGT. By comparing the results of strategy 1/2, 1/3, 2/2, 2/3, we can conclude that it is detrimental to perform the PGT refinement too frequently. \par

\begin{table}[!h]
\caption{Ablation studies of the hyper-parameter $k$. The best accuracy is highlighted in bold.}
\centering
\scriptsize
\begin{tabular}{cccccc}
\toprule  
$k$ & 1& 2& 3& 4& 5  \\
%\midrule  
\hline
mAP &53.21&  53.53& \textbf{53.99}& 53.63& 53.21\\

\bottomrule 
\end{tabular}
\end{table}

\subsection{Ablation study}
We conducted an ablation study to evaluate the effectiveness of each component of PGTRNet. We first investigate the influence of different settings of $k$ on the PGT generation. The effectiveness of considering more bounding boxes on PGT generation (PGT-G) and PGT refinement (PGT-R) are then discussed. Note that experiments on exploring the influence of the value of $k$ is not combined with the proposed PGT refinement. \par
As shown in Table 4, the mAP increases modestly from 53.21\% to 53.99\% as $k$ changes from 1 to 3. The main reason for this improvement is that an image may contain multiple instances from the same categories and a larger $k$ value allows algorithms to consider this case. This confirms the necessity of considering more bounding boxes when generating the PGT. Nonetheless, a different trend can be seen when $k$ continues to increase. The mAP reduces modestly from 53.99\% to 53.21\% when $k$ increases to 5. These reductions are caused by introducing more negative samples in PGT than positive samples. \par

Table 5 demonstrates the effectiveness of each component (PGT-G and PGT-R) of PGTRNet. PGT-G denotes the method of taking advantage of top three bounding boxes when generating the PGT. PGT-R is the PGT refinement strategy 2/3 with top two bounding boxes since we should ensure that the half of the $k$ is an integer on PGT refinement. It can be seen that PGT-G modestly boosts the mAP from 53.21\% to 53.99\%. PGT-R further improves the mAP by 1.30\% to 55.29\%. Combined with PGT-G, a significant improvement can be seen on PGT-R from 53.21\% to 55.29\% (+2.08\%) which achieves the state-of-the-art performance.

\begin{table}
\scriptsize
\caption{The ablation study results of each component (PGT-G and PGT-R) of PGTRNet on the test set of PASCAL VOC 2007 benchmark. }
\centering
\begin{tabular}{ccccc}
\toprule 
PGT-G &PGT-R &mAP\\
 \hline
- &- &53.21\\
$\surd$ &-  &53.99 \\
$\surd$ &$\surd$ &55.29\\
\bottomrule 
\end{tabular}
\end{table}

\section{Conclusion}
 Recent state-of-the-art WSOD approaches integrate the WSOD detector and fully-supervised detector. These methods often neglect the significance of the second-phase training, suffering from the insufficient learning problem.  We propose to consider more bounding boxes when creating the PGT. The low quality PGT generated by the WSOD detector deleteriously affects the performance of the fully supervised detector. To mitigate these problems, we propose a novel two-phase WSOD model PGTRNet which takes advantage of multiple bounding boxes to generate the PGT and gradually improves the quality of PGT. PGTRNet is simple yet effective without introducing extra learnable parameters. Experimental results and the ablation study demonstrate the superiority of PGTRNet which achieves the state-of-the-art performance on PASCAL VOC 2007 benchmark.

\vfill\pagebreak

\bibliographystyle{IEEEbib}
\bibliography{strings,refs}

\end{document}